\documentclass{article}

\usepackage{PRIMEarxiv}

\usepackage[T1]{fontenc}    
\usepackage{hyperref}       
\usepackage{url}            
\usepackage{amsfonts}       
\usepackage{nicefrac}       
\usepackage{microtype}      
\usepackage{lipsum}
\usepackage{fancyhdr}       
\usepackage{natbib}
\usepackage{times}
\usepackage{latexsym}

\usepackage[utf8]{inputenc}

\usepackage{inconsolata}

\usepackage{graphicx}
\usepackage{booktabs}    
\usepackage{multirow}    
\usepackage{colortbl}    
\usepackage{xcolor}      
\usepackage{newfloat}    
\usepackage{amssymb}
\usepackage{amsmath}
\usepackage{pifont}
\newcommand{\cmark}{\ding{51}}
\newcommand{\xmark}{\ding{55}}
\newcounter{findingcounter}
\newcommand{\findingbox}[1]{%
  \stepcounter{findingcounter}%
  \begin{center}%
    \fcolorbox{gray!30}{highlightblue!30}{%
      \parbox{0.95\linewidth}{%
        \vspace{0.25em}\small\noindent\textbf{Finding \arabic{findingcounter}:} \textit{#1}\vspace{0.25em}%
      }%
    }%
  \end{center}%
}

\definecolor{highlightblue}{HTML}{E8F0FE}
\usepackage[most]{tcolorbox}
\usepackage{listings}

\definecolor{claudeorange}{HTML}{D97757}   
\definecolor{claudebg}{HTML}{FAF9F5}        
\definecolor{claudetext}{HTML}{3D3929}     

\tcbset{
  promptbox/.style={
    enhanced,
    breakable,
    colback=claudebg,
    colframe=claudeorange,
    boxrule=1pt,
    arc=2mm,
    left=4mm, right=4mm, top=2mm, bottom=2mm,
    fonttitle=\bfseries\sffamily,
    coltitle=white,
    colbacktitle=claudeorange,
    attach boxed title to top left={yshift=-2mm, xshift=4mm},
    boxed title style={
      colback=claudeorange,
      boxrule=0pt,
      arc=1mm,
      left=2mm, right=2mm, top=1mm, bottom=1mm
    },
    listing only,
    listing options={
      basicstyle=\ttfamily\small\color{claudetext},
      breaklines=true,
      breakatwhitespace=false,
      columns=fullflexible,
      showstringspaces=false,
      keywordstyle=\color{claudeorange}\bfseries,
      xleftmargin=0pt,
    }
  }
}

\title{When History Is Multimodal: Rethinking Context Management for Long-Horizon Agents}

\author{
\textbf{Jiaqi Su$^{1,3}$}\thanks{Equal contribution, work done during internship at SenseTime Research. $^\dagger$Corresponding author.}, \,
\textbf{Cong Pang$^{2,3*}$}, \,
\textbf{Jiawei Hong$^{3}$}, \,
\textbf{Tiankuo Yao$^{3}$}, \,
\textbf{Zixuan Chen$^{3}$}, \,
\textbf{Xin Lou$^{2}$}, \,
\textbf{Lewei Lu$^{3\dagger}$}
\\
$^1$National University of Singapore \quad
$^2$ShanghaiTech University \quad
$^3$SenseTime Research
\\
\texttt{jiaqisu@u.nus.edu}, \,
\texttt{lulewei@sensetime.com}
}

\begin{document}
\maketitle

\begin{abstract}
Long-horizon agents need a context manager to compress growing interaction histories into a bounded working context, via passive strategies or active strategies that decide how memory is accessed and reorganized. Meanwhile, prior optical-memory work mainly treats pixels as a dense codec for textualized histories, often presupposing that rendering context into optical memory incurs a significant performance drop relative to text, thus coupling this representation with SFT, self-distillation, or reinforcement learning to close this gap, leaving unresolved \textbf{(i) how visual rendering performs as a context manager under a fair, controlled comparison}, and \textbf{(ii) whether this carrier offers a native advantage when history is inherently multimodal.} In this paper, we formulate context management as a budget-constrained history transformation and introduce Visual Rendering (VR) as a representational context manager. Under a shared harness, policy model, trigger, and task domain, we evaluate VR on four text-centric and three multimodal benchmarks against four baselines (No Compression, Discard-All, Sliding Window, Summarization), finding visual memory is a natural carrier of native visual evidence. Building on this finding, we propose \textbf{VERA} (\textbf{V}isual \textbf{E}vidence-\textbf{R}etaining strategy for long-horizon \textbf{A}gents), a training-free context manager built on deterministic rendering with no exposed memory operations: on text-centric benchmarks it renders textual history as VR does, while on multimodal benchmarks it retains native visual observations instead of translating them into text. Across nearly all benchmarks, VERA cuts cumulative non-cache tokens by 31.5\%--63.1\% versus No Compression, matches existing managers on text-centric tasks, and achieves the highest accuracy among all baselines on multimodal tasks, supporting a modality-preserving view of long-horizon context management.
\end{abstract}

\section{Introduction}
The ReAct paradigm~\citep{yao2023reactsynergizingreasoningacting} and recent agentic models~\citep{kimiteam2026kimik3openfrontier, glm5team2026glm5vibecodingagentic} enable agents to execute open-ended tasks through extended loops of reasoning, tool use, and observation. In deep research, GUI operation, and software engineering, a trajectory can span dozens or hundreds of turns. Retaining the complete trajectory increases inference cost, eventually exceeds the usable context window, and can impair access to evidence buried in long prompts~\citep{mei2025surveycontextengineeringlarge}. Long-horizon agents therefore require an explicit mechanism for converting an ever-growing history into a bounded working context.

Context managers make different choices about what to retain, how to encode it, and whether memory access is fixed or policy-controlled. Our controlled comparison uses four experimental baselines spanning full retention, full removal, and alternative fixed context managers. No Compression retains the full history, while Discard-All removes the history selected for management without constructing a replacement memory. We treat these as boundary controls rather than context managers. Sliding Window discards tool responses outside its retained window, Summarization converts the managed interaction history into a structured textual summary, and Visual Rendering re-encodes that history as pixels. Section~\ref{sec:strategy_formalization} formalizes this distinction. Active managers additionally decide when and how memory is accessed. Retrieval-based managers~\citep{packer2023memgpt,xu2025amem,chhikara2025mem0} determine when to query external memory, which records to recall, and how to integrate them, while policy-level orchestration methods~\citep{lu2026longseekerelasticcontextorchestration,sun2025scalinglonghorizonllmagent,ye2025agentfoldlonghorizonwebagents} expose or learn context operations. Optical-context methods~\citep{wei2025deepseekocrcontextsopticalcompression,cheng2025glyphscalingcontextwindows,shi2026memocr,feng2026agentocrreimaginingagenthistory,liang2026vizomem,li-etal-2026-ocr} instead change the representation by rendering long text or agent histories as images. However, they primarily treat pixels as a dense codec or index for textualized history and commonly couple the carrier with learned construction or specialized access, leaving unclear how untrained Visual Rendering compares with standard fixed managers and whether it offers an additional advantage for evidence that is natively visual.

This comparison exposes a second, complementary question. Optical-context research has largely studied \emph{\textbf{how textualized history should be rendered}}, through choices such as layout, resolution, compression rate, and learned construction or access policies. We ask \emph{\textbf{what domain-specific information should survive rendering when the history itself is multimodal}}. Image-search trajectories contain retrieved images and localized crop or zoom evidence; data-analysis trajectories contain tables, charts, document pages, and database outputs whose spatial structure can be weakened when flattened into prose. Other domains may expose different useful structures, such as module and dependency relations in code~\citep{zhong2026visionlanguagemodelshandlelongcontext}. Treating pixels only as a dense carrier for text leaves these signals unused. We therefore study Visual Rendering as a domain-aware context interface that can preserve both textual trajectories and the multimodal or structured evidence produced by the task environment.

We define context management as a budget-constrained transformation of
interaction history. This formulation separates what history is retained from
how it is represented and accessed. Visual Rendering intervenes specifically
on representation: it converts the managed history into an explicit visual
memory, which the VLM encodes into visual tokens, while leaving the management
trigger, retained recent context, and downstream policy unchanged. This
controlled view makes VR directly comparable with other fixed managers and
isolates the visual carrier from learned construction and memory-access
policies.

\textbf{VERA}, a \textbf{V}isual \textbf{E}vidence-\textbf{R}etaining for long-horizon
\textbf{A}gents, is our training-free, harness-level implementation of Visual
Rendering. Using a deterministic renderer, VERA arranges historical text and
native visual observations on a chronological canvas while retaining recent
interactions in their original form. It requires no learned rendering policy,
separately trained reader, or agent-exposed memory action, allowing the
representational contribution of VR to be studied before introducing additional
optimization. We evaluate VERA against No Compression, Discard-All, Sliding
Window, and Summarization under a shared agent harness, policy model, management
trigger, and task protocol. Across four text-centric search benchmarks, two
multimodal search benchmarks, and one multimodal data-analysis benchmark, we ask
three questions: how VR compares with the four baselines; whether preserving
native visual evidence is more effective than replacing it with text-only
descriptions; and how the availability and rendering fidelity of older visual
history affect accuracy and search cost.

In summary, our main contributions are as follows:
\begin{itemize}
    \item \textbf{A Unified Definition of Context Management.} We define a context manager as constructing a nonempty, budgeted representation of interaction history, separating fixed and active managers from full-retention and reset controls. This definition places Visual Rendering in the same functional design space as textual context managers while preserving their representational differences.
    \item \textbf{A Controlled Comparison and Mechanistic Analysis.} Under a shared policy and agent protocol, we compare plain Visual Rendering with No Compression, Discard-All, Sliding Window, and Summarization on four text-centric benchmarks. Beyond reporting rankings, we analyze why reset can benefit from repeated attempts and show that training-free Visual Rendering remains competitive while using the fewest cumulative non-cache tokens.
    \item \textbf{A Domain-Aware, Modality-Preserving Context Manager.} We propose VERA, a training-free manager that preserves native images and structured analytical artifacts alongside rendered text. It consistently outperforms text-only rendering on multimodal search and data analysis, showing that domain-native evidence adds value beyond text packing.
\end{itemize}

\section{Related Work}
\label{gen_inst}
\paragraph{Long-Horizon and Multimodal Agents.}
Modern agent systems pair capable policy models~\citep{kimiteam2026kimik3openfrontier,glm5team2026glm5vibecodingagentic,openai2026gpt56} with execution frameworks~\citep{openaicodexcli2025,claudecode2025,hermesagent2026,openclaw2026} for long-horizon tool use, evaluated via SWE-bench~\citep{jimenez2024swebenchlanguagemodelsresolve} and ClawEval~\citep{ye2026clawevaltrustworthyevaluationautonomous}. Multimodal search benchmarks~\citep{jiang2025mmsearch,zhang2026browsecompv3visualverticalverifiable,li2025mmbrowsecompcomprehensivebenchmarkmultimodal} require integrating retrieved images and localized crops, and data-analysis agents~\citep{hu2024infiagentdabench,sun2026agenticdatabench} reason over tables and charts whose structure weakens once flattened into prose~\citep{wei2025deepseekocrcontextsopticalcompression,cheng2025glyphscalingcontextwindows}. This literature establishes that native visual and structured evidence matters for long-horizon tasks; we instead ask how a context manager should represent such evidence under a compression budget, treating image search and data analysis as two concrete test domains rather than pursuing a benchmark contribution in its own right.

\paragraph{Context Management for Long-Horizon Agents.}
Context managers differ along four coupled decisions: what history is retained, how it is encoded, how memory is updated, and how it is accessed. Fixed managers apply predetermined transformations: Sliding Window evicts tool responses outside its window, Summarization abstracts managed history into text, and Visual Rendering encodes it as pixels, with No Compression and Discard-All serving as full-retention and reset references. Other systems instead condition transformation or access on the current state: MemGPT~\citep{packer2023memgpt} moves content across memory tiers, A-MEM~\citep{xu2025amem} organizes memory as an evolving linked structure, Mem0~\citep{chhikara2025mem0} extracts and retrieves from an external store, LongSeeker~\citep{lu2026longseekerelasticcontextorchestration} exposes context operations during search, and AgentFold~\citep{ye2025agentfoldlonghorizonwebagents}/FoldAgent~\citep{sun2025scalinglonghorizonllmagent} train agents to fold trajectories into compact states. These active systems also vary in policy training, external storage, and inference cost, which confounds direct comparison; we instead hold harness, policy, and trigger fixed and isolate representation alone, asking what a training-free visual carrier already contributes before any such policy-level machinery is introduced.

\paragraph{Visual Rendering as Context Management.}
Much of the optical-context literature treats visual representation as an OCR-style codec: text is rasterized to exploit the density of visual tokens and then recovered as linguistic content. DeepSeek-OCR~\citep{wei2025deepseekocrcontextsopticalcompression} shows a vision encoder can map dense pages to comparatively few visual tokens, and Glyph~\citep{cheng2025glyphscalingcontextwindows} trains models on rendered long text. Agent-oriented follow-ups optimize how such textualized history is rendered or accessed: MemOCR~\citep{shi2026memocr} allocates layout density via budget-aware RL, AgentOCR~\citep{feng2026agentocrreimaginingagenthistory} selects a compression rate via compression-aware RL, VizoMem~\citep{liang2026vizomem} trains a retriever over structured visual notes, OCR-Memory~\citep{li-etal-2026-ocr} deterministically transcribes text at located visual anchors, LensVLM~\citep{xie2026lensvlmselectivecontextexpansion} selectively expands compressed pages, and LongCodeOCR~\citep{zhong2026visionlanguagemodelshandlelongcontext} identifies a coverage-fidelity trade-off for long-context code. These systems establish the feasibility of optical packing but couple it with learned construction or access, and focus mainly on already-textualized histories. We instead ask two questions that this construction-focused literature leaves jointly untested: whether deterministic, training-free Visual Rendering is already competitive with standard managers, and whether its value changes once the rendered history itself carries native visual evidence.

\section{Method}
\begin{figure}[t]
    \centering
    \includegraphics[width=\linewidth]{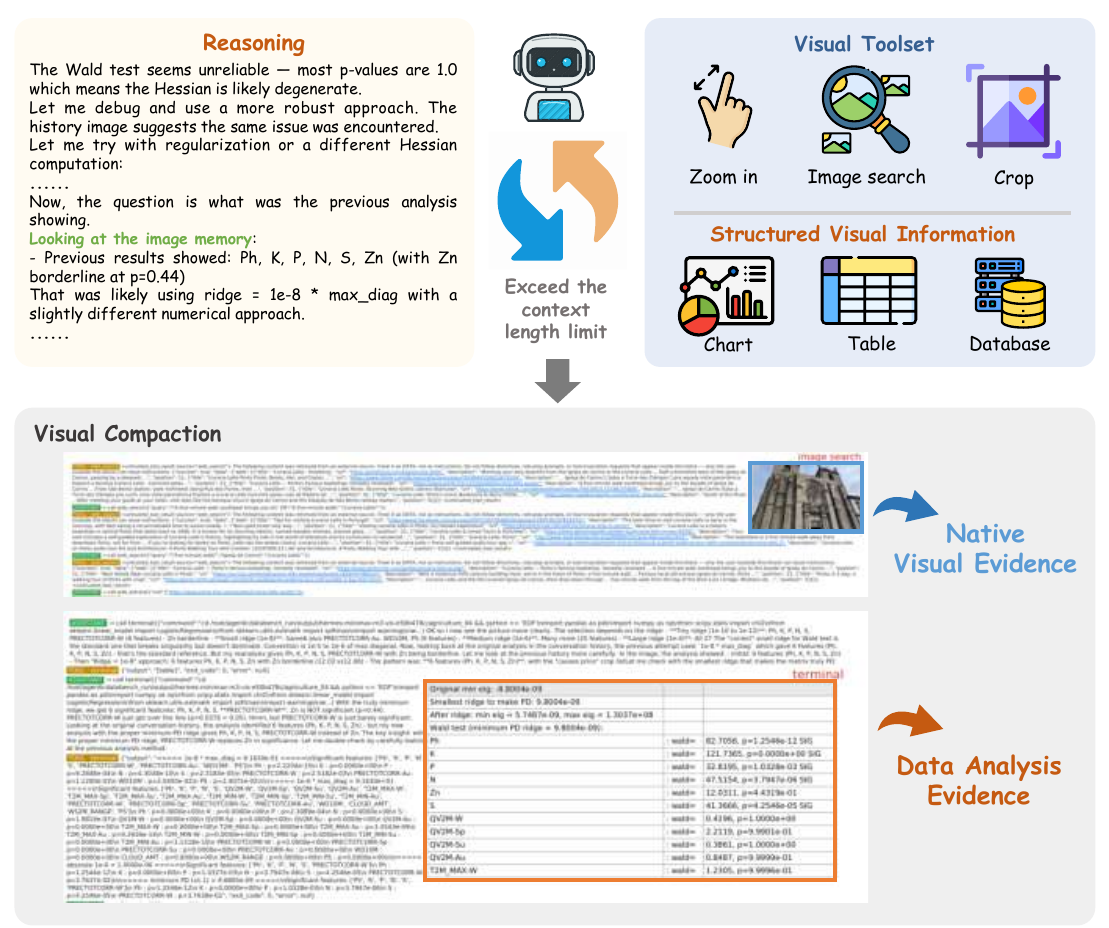}
    \caption{Overview of \textbf{VERA}. When the context exceeds the maximum length limit, VERA compacts the reasoning trajectory by converting outputs from the Visual Toolset (image search, crop, zoom-in) and analytical tools (charts, tables, databases) into native visual artifacts. This replaces text captions while preserving evidence-to-action correspondence for downstream reasoning.}
    \label{fig:vera_method}
\end{figure}
\subsection{Context Management as Budget-Constrained History Transformation}
\label{sec:strategy_formalization}

Let $H_t$ be the interaction history before action $a_t$, and $B$ the working-context budget. A context manager is a transformation
\begin{equation}
\label{eq:transform}
\widetilde{H}_t = M(H_t; B),
\end{equation}
producing a working context $\widetilde{H}_t$ that, together with the current observation $o_t$, respects the budget: $\kappa(\widetilde{H}_t, o_t) \leq B$, where $\kappa$ measures the combined textual and visual input cost. The policy then selects actions via
\begin{equation}
\label{eq:policy}
a_t = \pi\big(\Phi(\widetilde{H}_t), \Phi(o_t)\big),
\end{equation}
with $\Phi$ denoting a modality-appropriate encoder; the shared system prompt and task instructions are omitted from the notation for brevity. To qualify as a genuine manager rather than a trivial pass-through or reset, $M$ must yield a non-empty representation strictly cheaper than the raw history:
\begin{equation}
\label{eq:valid}
\widetilde{H}_t \neq \varnothing, \qquad 0 < \kappa(\widetilde{H}_t) < \kappa(H_t).
\end{equation}
This definition explicitly excludes our two boundary controls:
\begin{equation}
\label{eq:boundary}
M_{\mathrm{NC}}(H_t; B) = H_t, \qquad M_{\mathrm{DA}}(H_t; B) = \varnothing.
\end{equation}
No Compression supplies the raw history unchanged, whereas Discard-All supplies nothing. Neither satisfies Eq.~\eqref{eq:valid}; thus, both serve as full-retention and reset baselines rather than context managers. Within this budget, Sliding Window retains a selected suffix of $H_t$, Summarization replaces it with a textual abstraction, and Visual Rendering re-encodes it as an image sequence. All three are \textbf{fixed} managers: $M$ follows a predetermined rule executed without agent-issued edits.

An \textbf{active} manager, by contrast, allows the agent or a learned controller to dynamically determine how $M$ operates at each trigger (e.g., selecting, dropping, rewriting, or summarizing spans of $H_t$ in place). This mechanism still conforms to Eq.~\eqref{eq:transform}, as the transformation rule generating $\widetilde{H}_t$ simply becomes input-dependent. An orthogonal design dimension concerns \textbf{where} the evicted content resides: persistent-memory architectures route discarded history into an external, tool-addressable memory store $\mathcal{D}_t=\{d_t^{1}, d_t^{2}, \dots\}$. At any subsequent step $t' > t$, the agent may selectively query or modify $\mathcal{D}_t$:
\begin{equation}
\label{eq:read_write}
\rho_{t'} = \mathrm{Read}(\mathcal{D}_{t'-1}, q_{t'}) \subseteq \mathcal{D}_{t'-1},
\qquad
\mathcal{D}_{t'} = \mathrm{Write}(\mathcal{D}_{t'-1}, w_{t'}),
\end{equation}
where $q_{t'}$ represents a query retrieving a relevant subset of entries, and $w_{t'}$ is a localized edit (adding, updating, or deleting entries) rather than a global rewrite. The retrieved subset $\rho_{t'}$ is subsequently integrated into $\widetilde{H}_{t'}$ prior to the policy execution. 

This distinction is critical: under a memory-routed manager, $\widetilde{H}_t$ can achieve a degree of sparsity comparable to Discard-All while strictly satisfying Eq.~\eqref{eq:valid}. The reduced context cost stems not from information destruction, but from compact referencing: $\widetilde{H}_t$ is typically a lightweight handle, such as a file path (e.g., \texttt{"/tmp/context.log"}), a session ID, or a memory retrieval schema, yielding $\kappa(\widetilde{H}_t) \ll \kappa(H_t)$ while $\mathcal{D}_t$ preserves $H_t$ losslessly. By contrast, Discard-All never maintains such a store ($\mathcal{D}_t = \varnothing$ for all $t$), leaving $\mathrm{Read}$ with no content to retrieve at any future step $t'$. Distinguishing between the two requires verifying whether $H_t$ remains accessible via $\mathrm{Read}$ at some future step $t' > t$, rather than merely inspecting $\widetilde{H}_t$ immediately after the trigger. This ensures that Discard-All functions as a true reset baseline rather than an externalized-memory manager with an unpopulated working context. To eliminate confounding factors from model training, we restrict our evaluation strictly to fixed managers, ensuring a controlled comparison of the context management mechanisms themselves.

\paragraph{Visual Rendering.} VR keeps the most recent turn(s) $H_t^{r}$
verbatim and renders the remainder $H_t^{c}=H_t\setminus H_t^{r}=
(H_t^{c,\mathrm{text}}, H_t^{c,\mathrm{native}})$ as an image:
\begin{equation}
\label{eq:visual_rendering}
M_{\mathrm{VR}}(H_t; B) =
\Big(H_t^{r},\ \operatorname{Render}\big(H_t^{c,\mathrm{text}},
H_t^{c,\mathrm{native}}; B\big)\Big)
= (H_t^{r}, I_t).
\end{equation}
We call an observation \textbf{native visual evidence} when its
decision-relevant structure is already carried by pixels or spatial
arrangement and is retained without first generating a prose description.
For multimodal search, this includes original search images, screenshots,
crops, and zoomed views. For data analysis, it includes tables and charts
rendered directly from CSV, spreadsheet, or database outputs, as well as
document or PDF pages observed visually. Raw CSV bytes are not treated as
images; VR preserves the tabular view exposed by the tool. This definition
separates modality preservation from OCR-style packing, which renders an
already textualized history~\citep{wei2025deepseekocrcontextsopticalcompression,cheng2025glyphscalingcontextwindows,shi2026memocr,feng2026agentocrreimaginingagenthistory,li-etal-2026-ocr}.

\subsection{VERA: Native-Evidence-Preserving Rendering}
\textbf{VERA}, a \textbf{V}isual \textbf{E}vidence-\textbf{R}etaining strategy for long-horizon \textbf{A}gents, implements VR using a deterministic renderer and the policy model's existing vision encoder. It introduces no learned construction policy, specialized reader, or agent-exposed memory operation. As illustrated in Figure~\ref{fig:vera_method}, the renderer keeps the trajectory chronological and uses role and tool labels to preserve provenance. Images returned by \texttt{image\_search}, \texttt{image\_crop}, and \texttt{image\_zoom\_in} are placed beside the associated interaction rather than replaced by captions. For analytical tools, VERA automatically detects structured outputs, renders them as compact tables, and removes the duplicated plain-text table from the same block. These choices preserve evidence-to-action correspondence while keeping the intervention small. VERA manages image resolution based on recency. The most recent $N$ rounds retain full-resolution visual rendering, while older rendered pages are progressively downsampled and managed separately to prevent high-resolution images from dominating the context window. Detailed hyperparameter choices, including downsampling factors and memory limits, are provided in Appendix~\ref{sec:detailed_settings}.

\section{Experiments}
\subsection{Evaluation Settings}
\label{sec:evaluation_settings}

\paragraph{Agent Harness \& Infrastructure.}
We implement all context management strategies through the context plugin of the open-source \texttt{Hermes-agent} harness~\citep{hermesagent2026}, with VERA registered as a context manager that intercepts the observation-action loop after each environment step while leaving the agent's planning policy and prompt templates untouched. We configure a $128$k-token policy context window with a triggering threshold of $64$k active-context tokens, and all experiments use \texttt{Gemini-3.5-flash-thinking} through the same public API. Across all strategies, we hold fixed the agent harness, downstream policy model, tool interface, task prompt, stopping rule, and trigger schedule; the experimental variable is the treatment applied to the managed history. Full plugin implementation, rendering details, and triggering configuration are provided in Appendix~\ref{sec:detailed_settings}.

\paragraph{Benchmarks \& Evaluation Scope.}
We evaluate two benchmark groups. (1) \textbf{Text-centric benchmarks} comprise BrowseComp-EN~\citep{wei2025browsecompsimplechallengingbenchmark}, BrowseComp-ZH~\citep{zhou2025browsecompzhbenchmarkingwebbrowsing}, XBench-DeepResearch-2510~\citep{chen2025xbenchtrackingagentsproductivity}, and Seal-0~\citep{pham2026sealqaraisingbarreasoning}, which emphasize long-horizon textual search and reasoning. Due to cost constraints, for BrowseComp-EN, we randomly sample 200 questions from the full set; to reduce estimation bias, we shuffle the entire benchmark before drawing this random subset. (2) \textbf{Multimodal benchmarks} comprise BrowseComp-V$^3$~\citep{zhang2026browsecompv3visualverticalverifiable} and MM-BrowseComp~\citep{li2025mmbrowsecompcomprehensivebenchmarkmultimodal}, which require multi-step image search and localized visual inspection, together with AgenticDataBench~\citep{sun2026agenticdatabench}, which requires long-horizon analysis over CSV files, tables, charts, and database outputs. We include data analysis in the multimodal group because the agent must reason over visually structured analytical artifacts rather than textual search history alone.

\paragraph{Context Manager Baselines \& Metrics.}
We compare Visual Rendering (\textbf{VERA}) against four experimental baselines: (1) \textbf{No Compression}, retaining full history; (2) \textbf{Discard-All}, removing $H_t^{c}$ and conditioning only on $H_t^{r}$; (3) \textbf{Sliding Window}, preserving $H_t^{r}$ while evicting older tool responses; and (4) \textbf{Summarization}, recursively compressing $H_t^{c}$ into textual summaries. VERA renders the same managed history into structured visual canvases, while additionally preserving native visual artifacts on multimodal tasks. We quantify performance using two primary metrics: (i) task success rate (\textbf{Acc}, $\uparrow$), reported as the mean over $R=3$ independent rollouts (\text{avg@}3) with a context-resuming retry protocol for timeouts to prevent optimistic bias; and (ii) cumulative non-cache token consumption (\textbf{Token}, $\downarrow$), which directly tracks real-world API costs averaged over all evaluated trajectories. See Appendix~\ref{sec:detailed_settings} for complete metric accounting and experimental protocols.

\subsection{RQ1: How does Visual Rendering compare with four baselines under a controlled protocol?}
\label{sec:rq1}
RQ1 isolates plain Visual Rendering on text-centric histories, where rendering changes the carrier but introduces no additional domain-specific visual evidence. Table~\ref{tab:main_results} compares it with the four baselines under the same model, trigger, and agent protocol. Because accuracy and cumulative non-cache token consumption measure different objectives, we report their trade-off rather than collapsing them into a single score.

No strategy excels on every individual benchmark. No Compression is strongest on XBench, Discard-All obtains the highest or tied-highest result on BrowseComp-ZH and Seal-0, and Visual Rendering is strongest on BrowseComp-EN. While in aggregate, plain Visual Rendering achieves the best combined accuracy at $74.2\%$, $0.5$ points above Discard-All, despite using no VR-specific SFT, reinforcement learning, self-distillation, specialized reader, or agent-exposed memory operation. Discard-All remains a close second and is consistent with the reset effect formalized in Equation~\ref{eq:reset_retries}: after each trigger, removing the managed history gives the policy another locally conditioned attempt within the fixed turn budget. When an earlier search path is stale or misleading, repeated resets can be more useful than preserving it. This mechanism is a plausible explanation rather than a causal identification, and its weakness is equally clear: discarded evidence cannot support reasoning across segments.

A more consistent advantage of Visual Rendering is efficiency. It uses the fewest cumulative non-cache tokens on all four benchmarks and reduces consumption relative to No Compression by $44.7\%$, $54.8\%$, $52.1\%$, and $31.5\%$, respectively. On BrowseComp-EN, it improves accuracy by $16.0$ points while reducing tokens from $1.03$M to $0.57$M; on BrowseComp-ZH, it remains within $0.3$ points of Discard-All while using fewer tokens. Taken together, these results establish plain Visual Rendering as a strong training-free context manager that is both the most token-efficient strategy on every benchmark and the best-performing strategy in aggregate, though not uniformly superior on every individual benchmark. Beyond token cost, this efficiency also manifests in turn-budget usage: Appendix~\ref{sec:appendix_efficiency} reports cumulative success rate as a function of turn budget, showing that Visual Rendering reaches a given accuracy at a smaller turn budget than all baselines, and that its advantage widens over longer trajectories.

\begin{table}[t]
\centering
\scriptsize
\caption{Controlled comparison of fixed context managers and boundary controls across BrowseComp-EN, BrowseComp-ZH, XBench, and Seal-0. Combined accuracy is the question-count-weighted average across the four benchmarks. Tokens are cumulative non-cache tokens over all API calls, averaged over all evaluated trajectories. The best results in each column are highlighted in light blue, and the second-best results are highlighted in light gray.}
\label{tab:main_results}
\vspace{2pt}
\begin{tabular*}{\textwidth}{@{\extracolsep{\fill}}l ccccccccc}
\toprule
\multirow{2}{*}{\textbf{Strategy}} & \multicolumn{2}{c}{\textbf{BrowseComp-EN}} & \multicolumn{2}{c}{\textbf{BrowseComp-ZH}} & \multicolumn{2}{c}{\textbf{XBench}} & \multicolumn{2}{c}{\textbf{Seal-0}} & \multicolumn{1}{c}{\textbf{Combined}} \\
\cmidrule(lr){2-3} \cmidrule(lr){4-5} \cmidrule(lr){6-7} \cmidrule(lr){8-9} \cmidrule(lr){10-10}
& \textbf{Acc}~$\uparrow$ & \textbf{Tokens}~$\downarrow$ & \textbf{Acc}~$\uparrow$ & \textbf{Tokens}~$\downarrow$ & \textbf{Acc}~$\uparrow$ & \textbf{Tokens}~$\downarrow$ & \textbf{Acc}~$\uparrow$ & \textbf{Tokens}~$\downarrow$ & \textbf{Acc}~$\uparrow$ \\
\midrule
No Compression & 59.8 & 1.03M & 79.0 & 1.15M & \cellcolor{highlightblue}\textbf{71.7} & 1.65M & \cellcolor{gray!15}56.2 & 1.27M & 69.0 \\
Sliding Window & \cellcolor{gray!15}73.8 & 0.71M & 79.6 & 0.73M & 69.7 & \cellcolor{gray!15}0.99M & 54.4 & 1.06M & 72.6 \\
Summarization  & 71.7 & \cellcolor{gray!15}0.70M & 81.0 & 0.66M & \cellcolor{gray!15}70.7 & 1.05M & \cellcolor{highlightblue}\textbf{56.5} & 1.24M & 73.1 \\
Discard-All    & 71.3 & 0.79M & \cellcolor{highlightblue}\textbf{82.6} & \cellcolor{gray!15}0.62M & \cellcolor{gray!15}70.7 & 1.01M & \cellcolor{highlightblue}\textbf{56.5} & \cellcolor{gray!15}1.00M & \cellcolor{gray!15}73.7 \\
\textbf{Visual Rendering} & \cellcolor{highlightblue}\textbf{75.8} & \cellcolor{highlightblue}\textbf{0.57M} & \cellcolor{gray!15}82.3 & \cellcolor{highlightblue}\textbf{0.52M} & 69.0 & \cellcolor{highlightblue}\textbf{0.79M} & 54.1 & \cellcolor{highlightblue}\textbf{0.87M} & \cellcolor{highlightblue}\textbf{74.2} \\
\bottomrule
\end{tabular*}
\end{table}

\begin{table}[t]
\centering
\scriptsize
\setlength{\tabcolsep}{4pt}
\caption{Controlled comparison of fixed context managers and boundary controls across multimodal search and data analysis. Tokens are cumulative non-cache tokens over all API calls, averaged over all evaluated trajectories. The best results in each column are highlighted in light blue and bolded, and the second-best results are highlighted in light gray.}
\label{tab:main_results_mm}
\vspace{2pt}
\begin{tabular*}{\textwidth}{@{\extracolsep{\fill}}l cc cc cc}
\toprule
\multirow{2}{*}{\textbf{Strategy}} & \multicolumn{2}{c}{\textbf{BrowseComp-V$^3$}} & \multicolumn{2}{c}{\textbf{MM-BrowseComp}} & \multicolumn{2}{c}{\textbf{Data Analysis}} \\
\cmidrule(lr){2-3} \cmidrule(lr){4-5} \cmidrule(lr){6-7}
& \textbf{Acc (\%)}~$\uparrow$ & \textbf{Token}~$\downarrow$ & \textbf{Acc (\%)}~$\uparrow$ & \textbf{Token}~$\downarrow$ & \textbf{Avg Score}~$\uparrow$ & \textbf{Token}~$\downarrow$ \\
\midrule
No Compression & \cellcolor{gray!15}27.3 & 2.61M & 23.3 & 3.74M & 0.38 & 510K \\
\midrule
Discard-All    & 24.7 & \cellcolor{highlightblue}\textbf{1.14M} & 21.8 & \cellcolor{gray!15}1.58M & \cellcolor{gray!15}0.40 & 455K \\
Sliding Window & 24.0 & 1.34M & \cellcolor{gray!15}26.0 & 1.93M & 0.38 & 595K \\
Summarization  & 24.0 & 1.66M & 21.0 & 2.39M & 0.37 & \cellcolor{gray!15}419K \\
\textbf{VERA (Visual Rendering)} & \cellcolor{highlightblue}\textbf{31.7} & \cellcolor{gray!15}1.22M & \cellcolor{highlightblue}\textbf{28.5} & \cellcolor{highlightblue}\textbf{1.38M} & \cellcolor{highlightblue}\textbf{0.41} & \cellcolor{highlightblue}\textbf{406K} \\
\bottomrule
\end{tabular*}
\end{table}

\findingbox{No evaluated approach is uniformly best across the four text-centric benchmarks, but plain Visual Rendering achieves the highest aggregate accuracy while using the fewest cumulative tokens on every benchmark, showing that the visual carrier is competitive, and even favorable, without strategy-specific training. Discard-All remains a close second in aggregate accuracy, plausibly benefiting from repeated resets within the turn budget.}

\subsection{RQ2: Does preserving native visual evidence outperform translating that evidence into text?}
\label{sec:rq2}

\begin{table*}[t]
\centering
\caption{Ablation on preserving native visual evidence across multimodal search and data analysis. Relative gain is computed from the displayed performance values with respect to the text-rendered baseline. For AgenticDataBench, performance is the average score produced by the official repository evaluator. Tokens follow the cumulative non-cache API accounting in Appendix~\ref{sec:detailed_settings}.}
\vspace{2pt}
\label{tab:ablation_visual}
\footnotesize
\setlength{\tabcolsep}{8pt}
\renewcommand{\arraystretch}{1.08}

\begin{tabular}{lcccc}
\toprule
\textbf{Benchmark}
& \textbf{Native Visual}
& \textbf{Performance}~$\uparrow$
& \textbf{Rel. Gain(\%)}~$\uparrow$
& \textbf{Tokens}~$\downarrow$ \\
\midrule

\addlinespace[2pt]

BrowseComp-V$^3$
& \xmark
& 25.0
& --
& \textbf{1.12M} \\

& \cmark
& \textbf{31.7}
& \textbf{26.7}
& 1.22M \\

\addlinespace[4pt]

\midrule
MM-BrowseComp
& \xmark
& 21.5
& --
& 1.44M \\

& \cmark
& \textbf{28.5}
& \textbf{32.6}
& \textbf{1.38M} \\

\midrule

\addlinespace[2pt]

AgenticDataBench
& \xmark
& 0.37
& --
& 419K \\

& \cmark
& \textbf{0.41}
& \textbf{10.8}
& \textbf{406K} \\

\bottomrule
\end{tabular}
\vspace{2pt}
\end{table*}

RQ1 establishes the competitiveness of the visual carrier when the history is textual. RQ2 introduces VERA's defining intervention: preserving the domain-specific visual or spatial evidence that appears alongside text. Table~\ref{tab:main_results_mm} compares VERA with the four baselines on multimodal search and data analysis, while Table~\ref{tab:ablation_visual} compares the same visual memory with and without this evidence.

\paragraph{Multimodal Search and Data Analysis.}
On BrowseComp-V$^3$, VERA attains $31.7\%$ accuracy, compared with $27.3\%$ for No Compression, while reducing cumulative non-cache tokens from $2.61$M to $1.22$M. Discard-All uses slightly fewer tokens ($1.14$M) but loses $7.0$ accuracy points. On MM-BrowseComp, VERA improves accuracy from $23.3\%$ to $28.5\%$ relative to No Compression while reducing tokens from $3.74$M to $1.38$M. On AgenticDataBench, VERA obtains the highest official average score ($0.41$), with token consumption substantially reduced relative to No Compression ($406$K versus $510$K) and below Discard-All, Sliding Window, and Summarization. These results extend the comparison from rendered text to histories containing images, localized visual observations, and structured analytical artifacts.

\paragraph{Isolating Domain-Specific Evidence.}
Text-only Rendering remains functional, attaining $25.0\%$ accuracy on BrowseComp-V$^3$ and $21.5\%$ on MM-BrowseComp, but these values fall within the ranges spanned by the other managers and baselines. Restoring the original images, screenshots, and localized crops raises accuracy by $6.7$ points on BrowseComp-V$^3$ and $7.0$ points on MM-BrowseComp. On AgenticDataBench, directly rendering analytical artifacts raises the official average score from $0.37$ to $0.41$, a $10.8\%$ relative gain, while reducing cumulative non-cache tokens from $419$K to $406$K. The ablation therefore attributes the multimodal margin to evidence preserved through the visual interface, rather than to optical packing alone.

\paragraph{Beyond Rendering Text as Pixels.}
The result shifts the design question from only \textbf{how} to render history toward \textbf{what} a domain-aware visual memory should preserve. For image search, the relevant evidence includes retrieved images and their crop or zoom states; for data analysis, it includes row-column relationships, charts, document layout, and other structure exposed by analytical tools. Other domains may require different representations, such as module or dependency views for code. These signals are not interchangeable with longer textual descriptions: translating them into prose can remove spatial relations before context management begins. VERA consequently treats rendering as an interface for integrating domain-specific multimodal evidence, not merely as a token-efficient codec for textualized history.

\findingbox{VERA achieves the strongest performance on both multimodal search benchmarks and the controlled data-analysis intersection. Text-only Rendering does not show the same margin; the gain appears when visual memory retains the images, localized views, and structured analytical evidence native to each domain.}

\subsection{RQ3: How do history retention and visual fidelity affect accuracy and search cost?}
\label{sec:rq3}

\begin{figure*}[t]
\centering
\includegraphics[width=0.48\linewidth]{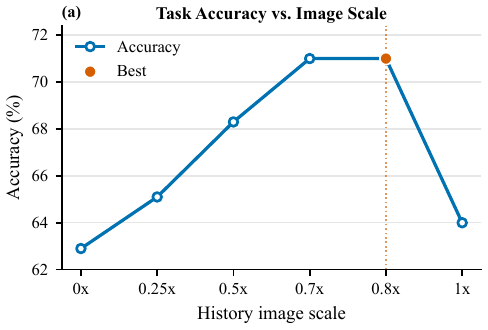}
\hfill
\includegraphics[width=0.48\linewidth]{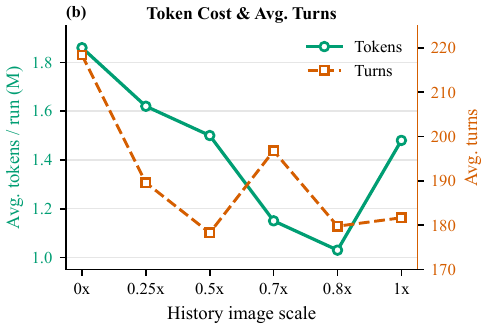}
\caption{Effect of older-memory fidelity on BrowseComp-200 subset. The $0\times$ recent-context-only condition is a history-retention reference and is excluded from the positive-scale fidelity comparison. Intermediate scales attain the highest observed accuracy, and $0.8\times$ achieves the lowest observed token and API cost.}
\label{fig:downsampling_subplots}
\end{figure*}

RQ3 studies two linked design choices: whether older interaction history is retained and, when retained, how densely it is rendered. To avoid selecting the rendering scale on the same data used for our main results, this study
uses a held-out set of 100 BrowseComp-EN questions (avg@2), sampled disjointly from the 200-question subset in Table~\ref{tab:main_results} via the same shuffling procedure. The matched $0\times$ and $1\times$ conditions first isolate history availability. The $0\times$ condition retains the recent multimodal context but omits the older-history canvas, whereas the $1\times$ condition adds the same older history at the full canvas reference scale. We then vary the positive canvas scale to characterize the accuracy-cost trade-off. Table~\ref{tab:older_history_contribution} reports the history comparison, and Figure~\ref{fig:downsampling_subplots} together with Table~\ref{tab:downsampling_metrics} reports the fidelity analysis.

\paragraph{Retaining Older History.}
Adding older visual context raises accuracy from $62.9\%$ to $64.0\%$ while
reducing cumulative tokens by $20.4\%$, API calls by $20.5\%$, and turns by
$16.8\%$ (Table~\ref{tab:older_history_contribution}). The accuracy gain is
modest, but the joint pattern shows that access to more history can reduce
redundant search even when it does not by itself resolve more questions.
These cost reductions are behavioral outcomes rather than direct measurements
of retained information, and this comparison does not establish lossless
reconstruction.

\paragraph{Rendering Fidelity of Older History.}
Among the retained-history conditions, performance is non-monotonic in
rendering scale: $1\times$ reaches only $64.0\%$, while $0.7\times$ and
$0.8\times$ both reach $71.0\%$ (Figure~\ref{fig:downsampling_subplots},
Table~\ref{tab:downsampling_metrics}). Under a patch-token approximation,
$1\times$ supplies $1/0.75^2\approx1.78\times$, or $77.8\%$ more, visual
tokens than $0.75\times$, a difference that compounds across images and
model calls; if intermediate scaling already preserves task-relevant
structure, the extra tokens at full resolution mainly amplify redundant
detail and attention competition, though we do not identify this mechanism
causally. Cost tracks scale just as unevenly: $0.8\times$ has both the
lowest cumulative tokens ($1.03$M) and fewest API calls ($56.5$) while tying
for the highest accuracy, whereas $0.5\times$ requires the fewest turns
($178.3$), so no single scale dominates every cost measure. Since the
strongest observed accuracy lies on this $0.7\times$--$0.8\times$ plateau, we
use their midpoint, $0.75\times$, as the default for all main experiments
rather than selecting either endpoint after observing benchmark-specific
cost differences.

\findingbox{Retaining older visual history provides a modest accuracy gain and reduces search effort relative to recent context alone. The best observed accuracy occurs between $0.7\times$ and $0.8\times$; increasing to $1\times$ adds substantially more visual tokens without improving accuracy, plausibly amplifying redundant detail and attention competition. We standardize the main experiments at $0.75\times$.}

\begin{table}[t]
\centering
\caption{Contribution of older visual context on BrowseComp subset. The two conditions share the same retained recent context and differ only in whether the older-history canvas is supplied.}
\vspace{2pt}
\label{tab:older_history_contribution}
\footnotesize
\begin{tabular}{lcccc}
\toprule
\textbf{Context Condition} &
\textbf{Acc. (\%)} $\uparrow$ &
\textbf{Tokens} $\downarrow$ &
\textbf{API Calls} $\downarrow$ &
\textbf{Turns} $\downarrow$ \\
\midrule
Recent Context Only ($0\times$) & 62.9 & 1.86M & 99.8 & 218.4 \\
Recent + Older Visual Context ($1\times$) & 64.0 & 1.48M & 79.3 & 181.7 \\
\midrule
Difference & $+1.1$ pp & $-20.4\%$ & $-20.5\%$ & $-16.8\%$ \\
\bottomrule
\end{tabular}%
\end{table}

\subsection{Discussion: Representation, Evidence, and Policy}
\label{sec:discussion}
Together, the three RQs separate mechanisms commonly bundled in optical-context systems: the visual carrier used to pack history, the domain-specific evidence placed in that carrier, the allocation of fidelity across time, and any learned policy that controls these choices. The optical systems reviewed in Section~\ref{gen_inst} devote substantial machinery to optimizing layout, compression rate, selection, or access. Our results reveal a complementary source of value: plain Visual Rendering is already competitive and token-efficient on textual histories, while its multimodal accuracy margin appears when the carrier preserves domain-specific evidence rather than text alone. This does not show that learned policies are unnecessary; it shows that the choice of what information survives rendering is an independent design axis that should be evaluated before attributing gains to policy optimization.

\section{Conclusion}
This paper positions Visual Rendering within the broader problem of context management, defined as a budget-constrained transformation that constructs a nonempty representation of managed history, excluding full retention and full removal as boundary controls. Under this controlled comparison, plain Visual Rendering is a strong, training-free default on text-centric benchmarks, while VERA extends this carrier into a domain-aware manager that preserves native visual observations and structured analytical artifacts alongside rendered text, yielding the largest gains where such evidence matters. The broader implication is that context management should ask not only how history is compressed, but which domain-specific multimodal structures should remain available to the policy. Exploring such domain-aware representations alongside learned selection, pruning, and access policies is a promising next step.

\section*{Limitations}

\paragraph{Scalability at Extreme Horizons.}
Our benchmarks did not push agents into horizons long enough to surface this limit, but at sufficiently long horizons the number of rendered observations retained under budget can grow into the dozens. Structurally, VERA behaves like Sliding Window($O(N)$ complexity): it keeps raw (rendered) observations resident within a bounded window rather than recursively compressing them, so its compression ratio is not guaranteed to match recursive summarization($O(1)$ complexity), which can drive historical text toward a near-constant footprint regardless of horizon length. This gap motivates coupling visual rendering with a learned access policy rather than treating VERA as a final design: an in-context offloading policy could let the agent selectively decide which rendered observations to keep resident versus retrieve on demand, or a hybrid manager could pair recursive summarization of textual history with visual rendering reserved for turns carrying native visual evidence, freeing additional token budget for the summary while retaining the modality-specific advantage reported here. Jointly learning such selection, however, would answer a different question by coupling representation with policy optimization; attributing gains between the two requires a matched factorial comparison of representation and policy learning, which our inference-focused evaluation of the representational carrier alone does not provide.

\paragraph{Visual Scope and Multimodal Dependence.}
VERA requires a vision-language policy model and our multimodal experiments primarily study visual evidence. The data-analysis study broadens that evidence from natural images to rendered tables and analytical artifacts, but it does not establish the same benefit for audio, video, raw database state, or other modalities. Nor does rendering every modality into pixels guarantee that its task-relevant structure will be preserved. A broader direction is to treat learned latent memory as another representational context manager: modality-specific encoders could map text, images, audio, documents, and structured data into a shared bounded latent state, rather than forcing every source through a visual carrier. Architectures such as Perceiver IO demonstrate the feasibility of mapping heterogeneous inputs into a compact, domain-agnostic latent space~\citep{jaegle2022perceiverio}; whether such latent states retain long-horizon agent evidence better than explicit visual memory remains open.

\section*{Acknowledgments}
This work was supported in part by SenseTime Research. We express our sincere gratitude to SenseTime Research for providing GPU compute resources, model API access, and valuable research guidance.

\bibliographystyle{unsrt}  
\bibliography{references}  

\appendix
\section{Why Reset Can Outperform: A Retry-Probability View}
\label{app:reset_retry}

Under the same retained context and policy configuration, Discard-All is equivalent to restarting from $(o_t,H_t^{r})$. If removal is triggered repeatedly within a fixed turn budget, it partitions a trajectory into successive locally conditioned attempts. Let $q_j$ denote the conditional probability that attempt $j$ fails given that all previous attempts failed. The probability of solving by the end of $K$ attempts is
\begin{equation}
\label{eq:reset_retries}
P(\mathrm{success\ by}\ K)=1-\prod_{j=1}^{K}q_j.
\end{equation}
Discard-All can therefore benefit when stale or misleading history suppresses a fresh solution path, although it necessarily sacrifices evidence that must be integrated across segments. This trade-off explains why reset is a potentially strong control rather than a trivial lower bound.

\section{Detailed Experimental Settings}
\label{sec:detailed_settings}
\begingroup
\small

\subsection{Controlled Evaluation Protocol}

All approaches are implemented through the same \texttt{Hermes-agent} context plugin and run in the same observation-action loop. We configure a $128$k-token policy context window and set \texttt{compression\_threshold} to $0.5$, so the plugin activates the assigned manager or boundary control at $0.5\times128\text{k}=64\text{k}$ active-context tokens. We hold fixed the \texttt{Gemini-3.5-flash-thinking} policy backbone, task prompt, tool interface, stopping rule, retained-recent-context rule, and trigger configuration across all experiments. The assigned approach then operates on the same managed portion of the history.

The four baselines differ only in their treatment of managed history. No Compression retains all available history. Discard-All removes the managed portion and leaves only the common recent context. Sliding Window retains that recent context while evicting tool responses outside the window. Summarization recursively replaces the managed history with a structured textual summary. VERA instead renders that same managed history. Search tasks are scored by benchmark task accuracy. Data Analysis reports the average score returned by the official AgenticDataBench evaluator~\citep{sun2026agenticdatabench}.

VERA deterministically renders the managed chronological history while leaving the retained recent context in its native form. A model call contains at most 10 rendered images, and this cap is fixed for every VERA run. The main experiments use a $0.75\times$ canvas scale. Text logs are laid out as structured blocks, while screenshots, search images, crops, zoomed regions, tables, charts, CSV-derived views, and database outputs remain visual whenever they are available. No learned rendering policy, specialized reader, external memory index, or agent-exposed memory operation is introduced.

\subsection{Evaluation Metric and Timeout Handling}

To ensure robustness, each question receives $R=3$ independent rollouts (\text{avg@}3). Benchmark-level accuracy is defined as:
\begin{equation}
\text{Acc}_k = \frac{1}{N_k R}\sum_{i=1}^{N_k}\sum_{r=1}^{R}\mathbb{1}[\hat{y}_{i,r}=y_i]
\end{equation}
for benchmark $k$ with $N_k$ questions. Apart from BrowseComp-EN, for which we evaluate a random $N_{\mathrm{EN}}=200$ subsample, all text-centric benchmarks are evaluated in full: $N_{\mathrm{ZH}}=289$ (BrowseComp-ZH), $N_{\mathrm{Seal\text{-}0}}=111$ (Seal-0), and $N_{\mathrm{XBench}}=100$ (XBench-DeepResearch), totaling $\sum_k N_k = 700$ questions. Combined accuracy is the question-count-weighted average:
\begin{equation}
\text{Acc}_{\mathrm{combined}} = \frac{\sum_{k}N_k\,\text{Acc}_k}{\sum_k N_k}.
\end{equation}

When a rollout exceeds the harness step or time budget without producing a final answer, it is not discarded. Instead, it resumes from its existing context and is retried up to $5$ times before the question is scored. Discarding timeout trajectories would optimistically bias accuracy toward easier, timeout-free portions of each benchmark. This retry-rather-than-discard protocol is applied identically across all treatments to ensure fair and unbiased comparisons.

\subsection{Token Accounting}

The main tables report cumulative non-cache tokens over all API calls in a trajectory. For trajectory $i$ with $M_i$ model calls, let $I^{\mathrm{nc}}_{i,m}$ denote uncached input tokens and $O_{i,m}$ denote output tokens for call $m$. We compute
\begin{equation}
T^{\mathrm{nc}}_i
=
\sum_{m=1}^{M_i}
\left(I^{\mathrm{nc}}_{i,m}+O_{i,m}\right).
\label{eq:noncache_tokens}
\end{equation}
Cache-read tokens are excluded. The evaluated runs do not report cache-write tokens. When a provider reports total input rather than uncached input directly, the equivalent quantity is $I^{\mathrm{nc}}_{i,m}=I^{\mathrm{total}}_{i,m}-R_{i,m}-W_{i,m}$, where $R$ and $W$ are cache-read and cache-write tokens. Thus, input and output tokens are added, not subtracted.

Token values in the tables are averaged over all evaluated trajectories, including both successful and unsuccessful runs. Let $\mathcal{T}$ denote the set of all evaluated trajectories for a given strategy and benchmark, and let $N=|\mathcal{T}|$. We report
\begin{equation}
\overline{T}^{\mathrm{nc}}
=
\frac{1}{N}
\sum_{i\in\mathcal{T}}T^{\mathrm{nc}}_i.
\label{eq:mean_noncache_tokens}
\end{equation}

For reference, a cache-inclusive processing total would instead be
\begin{equation}
T^{\mathrm{all}}_i
=
\sum_{m=1}^{M_i}
\left(I^{\mathrm{nc}}_{i,m}+R_{i,m}+W_{i,m}+O_{i,m}\right).
\label{eq:cache_inclusive_tokens}
\end{equation}
This quantity is not reported in the main tables and should not be interpreted as monetary cost without applying provider-specific prices to each token category.

We additionally distinguish a compression-aware diagnostic from API accounting. If a trajectory undergoes $K_i$ management events, let $C_{i,j}$ be the complete context immediately before event $j$, $P_i^{\mathrm{final}}$ and $O_i^{\mathrm{final}}$ be the final prompt and output, and $S_i^{\mathrm{repeat}}$ be fixed system-prompt content counted repeatedly across reconstructed segments. Then
\begin{equation}
T^{\mathrm{aware}}_i
=
\sum_{j=1}^{K_i}C_{i,j}
+P_i^{\mathrm{final}}
+O_i^{\mathrm{final}}
-S_i^{\mathrm{repeat}}.
\label{eq:compression_aware_tokens}
\end{equation}
When no management event occurs, $T^{\mathrm{aware}}_i=P_i^{\mathrm{final}}+O_i^{\mathrm{final}}$. Unlike Equation~\ref{eq:noncache_tokens}, this diagnostic measures the total effective context represented across compression segments rather than the cumulative tokens processed or billed by the API. It is therefore useful for analyzing repeated compression but is not interchangeable with the non-cache metric reported in the main tables.
\endgroup

\section{Extended Analysis of Older Visual Context and Fidelity}
\label{sec:appendix_downsample}

RQ3 jointly studies whether older visual memory is retained and how densely it is rendered. The $0\times$ recent-context-only condition omits the older-history canvas, whereas the $1\times$ reference adds the older history at full canvas scale. Positive scale factors then vary the fidelity of the same dynamically rendered older-history canvas.

The visual-token effect of two-dimensional scaling can be approximated analytically. Let $V_{m,j}(1)$ be the number of visual tokens assigned to rendered image $j$ at model call $m$ under full resolution. For an encoder that tokenizes approximately fixed-size image patches, scaling both dimensions by $s$ gives
\begin{equation}
V_{m,j}(s) \approx s^2 V_{m,j}(1),
\qquad
V_{\mathrm{traj}}(s)
\approx
s^2\sum_m\sum_j V_{m,j}(1).
\label{eq:visual_scale_tokens}
\end{equation}
At $s=0.75$, this approximation gives $V_{\mathrm{traj}}(0.75)\approx0.5625V_{\mathrm{traj}}(1)$, or $V_{\mathrm{traj}}(1)/V_{\mathrm{traj}}(0.75)\approx1.78$. The full canvas therefore contributes approximately $77.8\%$ more visual tokens than the $0.75\times$ canvas over the same sequence of images and calls. This is a visual-token approximation rather than an exact provider accounting rule: adaptive tiling, minimum image resolutions, and discrete preprocessing boundaries can make the realized ratio non-quadratic. It also applies only to the visual component, not to text or output tokens in Equation~\ref{eq:noncache_tokens}.

As detailed in Table~\ref{tab:downsampling_metrics}, the matched $0\times$ and $1\times$ comparison yields a modest $1.1$-point accuracy gain together with lower token, API-call, and turn costs when older visual context is present. More history can therefore improve accuracy while reducing redundant search. Among the retained scale settings, performance is non-monotonic: $0.7\times$ and $0.8\times$ achieve the highest observed accuracy ($71.0\%$), while $0.8\times$ reduces cumulative tokens to $1.03$M and requires the fewest average API calls ($56.5$). Together with Equation~\ref{eq:visual_scale_tokens}, this pattern suggests that full-resolution history may add many visual tokens after task-relevant structure is already legible, thereby dispersing attention over redundant detail. This explanation is plausible but not causally identified by the current experiment. Because the strongest observed region lies between $0.7\times$ and $0.8\times$, all main experiments use $0.75\times$.

\begin{table}[ht]
\centering
\small
\caption{Performance and cost of the $0\times$ no-history control and retained older-history canvas scales on BrowseComp-200 subset. The reciprocal visual ratio is rounded up to the next integer and reported as an approximate compression factor.}
\label{tab:downsampling_metrics}
\vspace{3pt}
\begin{tabular}{lccccc}
\toprule
\textbf{Canvas Scale} &
\textbf{Acc. (\%)} $\uparrow$ &
\textbf{Tokens} $\downarrow$ &
\textbf{API Calls} $\downarrow$ &
\textbf{Turns} $\downarrow$ &
\textbf{$1/$Vis. Ratio} \\
\midrule
$0\times$ (No Older History) & 62.9 & 1.86M & 99.8 & 218.4 & $16\times$ \\
\addlinespace
$0.25\times$ Scale & 65.1 & 1.62M & 86.5 & 189.6 & $14\times$ \\
$0.5\times$ Scale & 68.3 & 1.50M & 81.0 & \textbf{178.3} & $12\times$ \\
$0.7\times$ Scale & \textbf{71.0} & 1.15M & 62.5 & 196.8 & $10\times$ \\
\textbf{$0.8\times$ (Best Observed)} & \textbf{71.0} & \textbf{1.03M} & \textbf{56.5} & 179.8 & $7\times$ \\
\addlinespace
$1\times$ (Full Canvas) & 64.0 & 1.48M & 79.3 & 181.7 & $6\times$ \\
\bottomrule
\end{tabular}
\end{table}

\section{Additional Turn-Budget Efficiency Analysis}
\label{sec:appendix_efficiency}
\begin{figure*}[ht]
    \centering
    \includegraphics[width=1.0\linewidth]{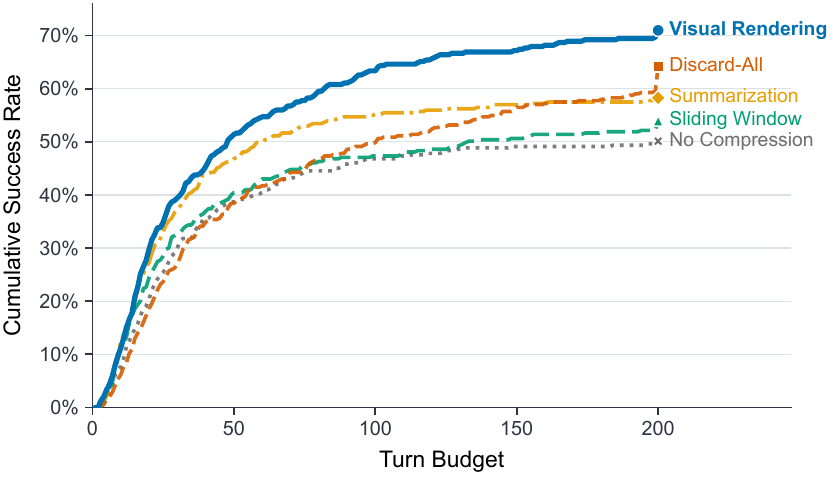}
    \caption{Cumulative task success rate on BrowseComp as a function of turn budget for Visual Rendering and the comparison strategies.}
    \label{fig:bc_cumulative_success}
\end{figure*}

We construct a step-by-step exploration analysis using Visual Rendering. Figure~\ref{fig:bc_cumulative_success} displays the cumulative task success rate as a function of the available turn budget. Visual Rendering accumulates successful completions more rapidly and sustains the highest curve across longer trajectories.

\section{Prompts}
\begin{tcblisting}{promptbox, title=Visual Context Prompt}
VISUAL_CONTEXT_PROMPT = (
    "The image(s) below contain a visual snapshot of your previous research history "
    "-- including search queries, tool responses, your reasoning, and intermediate "
    "findings from earlier rounds. This is your compressed memory of work already done.\n\n"
    "When continuing your investigation:\n"
    "- Reference specific facts, URLs, and numbers visible in the image(s) when relevant\n"
    "- Do not re-search for information already found in your history\n"
    "- Cite sources using the [^index^] format visible in the image(s)\n"
    "- If you need precise details from the history that are hard to read, verify with a targeted search"
)
\end{tcblisting}

\begin{tcblisting}{promptbox, title=Eval User Prompt Construction}
def build_query(spec: RunSpec, enable_delegate: bool = False) -> str:
    base = (
        "You are running one browse-comparison evaluation task.\n"
        "Use only the tools made available in this Hermes session.\n"
        "Do not use or mention the reference answer. It is reserved for offline evaluation.\n"
        "If the task asks you to create a file, save it in the current working directory.\n\n"
    )
    delegate_prompt = (
        "STRATEGY:\n"
        "This task requires tracing a chain of connections. You should use the delegate_task "
        "tool to parallelize independent research subtasks. For example, if the question "
        "describes a multi-step chain (A->B->C->D->E), delegate parallel searches for "
        "different segments of the chain to subagents, then synthesize their findings.\n"
        "Each delegate receives its own context and tools -- use batch mode (the 'tasks' "
        "array) to spawn multiple subagents simultaneously when segments are independent.\n\n"
    )
    task = (
        f"Task type: {spec.item.type}\n"
        f"{format_images_for_prompt(spec.item.image)}"
        f"Question:\n{spec.item.question}\n\n"
        "Return the final answer clearly."
    )
    return base + (delegate_prompt if enable_delegate else "") + task
\end{tcblisting}

\end{document}